\documentclass[12pt]{article}
\usepackage{amsmath}
\usepackage{graphicx,psfrag,epsf}
\usepackage{enumerate}
\usepackage{natbib}
\usepackage{natbib}
\usepackage{url}

\newcommand{\blind}{0} 

\def\spacingset#1{\renewcommand{\baselinestretch}%
{#1}\small\normalsize} \spacingset{1}

\usepackage{amssymb}
\usepackage{algorithm}
\usepackage{algorithmic}
\usepackage{array}
\usepackage{amsthm}
\usepackage{subfig}
\usepackage{multirow}   
\usepackage{color}
\usepackage[normalem]{ulem}
 \usepackage{indentfirst}
 \usepackage{times}
\usepackage{epsfig}
\usepackage{stmaryrd}
\usepackage{diagbox}
\usepackage{comment}
\usepackage[normalem]{ulem}
\usepackage{slashbox}
\usepackage{float}

\newtheorem{definition}{Definition}
\theoremstyle{remark}

\newtheorem{rem}{Remark}

\usepackage{hyperref}
\usepackage{color}
\usepackage[normalem]{ulem}
\newcounter{ToDo}
\newcounter{gaocomm}
\newcounter{Note}
\definecolor{blue-violet}{rgb}{0.54, 0.17, 0.89}
\definecolor{mygreen}{rgb}{0.0, 0.5, 0.0}
\definecolor{awesome}{rgb}{1.0, 0.13, 0.32}
\definecolor{bostonuniversityred}{rgb}{0.8, 0.0, 0.0}

\begin{document}

\date{}
\bibliographystyle{apalike}

\def\spacingset#1{\renewcommand{\baselinestretch}%
{#1}\small\normalsize} \spacingset{1}



\if0\blind
{
  \title{\bf Manifold optimization Assisted Gaussian Variational Approximation\thanks{The project is supported by the University of Sydney Business School Pilot Research Project Grant and Australian Research Council (ARC) Discovery Project, Grant DP200103015.}}
  \author{Bingxin Zhou,
    Junbin Gao, 
    Minh-Ngoc Tran
    and
    Richard Gerlach \\
    The Discipline of Business Analytics, The University of Sydney Business School\\
    The University of Sydney, NSW 2006, Australia\\
    bzho3923@uni.sydney.edu.au\\
    \{junbin.gao, minh-ngoc.tran, richard.gerlach\}@sydney.edu.au}
  \maketitle
} \fi

\if1\blind
{
  \bigskip
  \bigskip
  \bigskip
  \begin{center}
    {\LARGE\bf Title}
\end{center}
  \medskip
} \fi

\bigskip

\begin{abstract}
Gaussian variational approximation is a popular methodology to approximate posterior distributions in Bayesian inference especially in high dimensional and large data settings. To control the computational cost while being able to capture the correlations among the variables, the low rank plus diagonal structure was introduced in the previous literature for the Gaussian covariance matrix. For a specific Bayesian learning task, the uniqueness of the solution is usually ensured by imposing stringent constraints on the parameterized covariance matrix, which could break down during the optimization process. In this paper, we consider two special covariance structures by applying the Stiefel manifold and Grassmann manifold constraints, to address the optimization difficulty in such factorization architectures. To speed up the updating process with minimum hyperparameter-tuning efforts, we design two new schemes of Riemannian stochastic gradient descent methods and compare them with other existing methods of optimizing on manifolds. In addition to fixing the identification issue, results from both simulation and empirical experiments prove the ability of the proposed methods of obtaining competitive accuracy and comparable converge speed in both high-dimensional and large-scale learning tasks.
\end{abstract}

\noindent%

\newpage
\spacingset{1.5} 

\section{Introduction}
Variational inference is a well-established approach to approximate intractable posterior distributions in Bayesian statistics \citep{fox2012tutorial,blei2017variational,zhang2018advances}. Compared to sampling-based Monte Carlo algorithms \citep{gamerman2006markov}, this optimization-based method is computationally efficient and scalable to large datasets. Variational inference turns the Bayesian inference problem into an optimization problem that approximates the posterior distribution by the closest member from a family of tractable distributions such as Gaussians.
The optimal variational approximation is often solved by stochastic gradient descent algorithms \citep{hoffman2013stochastic}, which are able to handle massive datasets. 

Successful estimation of the posterior relies on determining an appropriate variational family. Typically, the multivariate Gaussian family is often used as it provides notable tractability and simplicity \citep{opper2009variational,ormerod2012gaussian}.
However, estimating a full covariance matrix in Gaussian variational approximation is generally challenging, since the number of matrix elements grows quadratically with the number of latent variables and the computational cost quickly becomes too expensive.  Consequently, special structures are applied to the covariance matrix for scalability. For example, the mean-field variational family assumes a diagonal structure for the covariance matrix \citep{xing2002generalized,gershman2012nonparametric}.
However, this assumption of independence is inappropriate to many complicated data structures and tends to underestimate the variance in practice \citep{blei2017variational}. An alternative is to factorize the covariance matrix using a low-dimensional representation. Ideally, such low-dimensional representation should be simple enough to maintain a tractable approximation, but expressive enough to capture well the posterior correlations. Various low-dimensional representation structures have been considered so far, where one common choice is the low-rank plus diagonal (LR+D) decomposition, see \cite{nickisch2009convex,seeger2010gaussian,guo2016boosting,miller2016variational}. In particular, \cite{ong2018gaussian} proposed a Gaussian variational approximation method with the factorized covariance structure (\textsc{VAFC}) via the LR+D representation. Much of our paper will build on the VAFC method. 

Originated from the factor analysis model in statistics, the LR+D decomposition requires restrictions to guarantee global identification of the factors to be estimated. The topic has been studied for a long time, see \cite{bekker1986note,grayson1994identification,grzebyk2004identification}, to mention but a few. For example, the restrictions in \cite{bai2013principal} require the underlying factor matrix to be orthogonal lower trapezoidal. 
In the Euclidean space, imposing extra constraints can result in a non-convex optimization problem because of the changes in the feasible space. \cite{ong2018gaussian} retained Euclidean optimization by ignoring the full-rank restriction on the factor matrix. 
This might lead to an inefficient optimization procedure and make it likely trapped into a bad local mode because the factor matrix is not uniquely determined. 

Evidence shows that a constrained optimization problem with non-Euclidean nature can be transformed into a geodesically convex problem in the Riemannian space \citep{edelman1998geometry,boumal2014optimization,cherian2016riemannian,ferreira2018gradient}. Various manifolds have been applied for geometry-aware learning tasks \citep{huo2007survey,fiori2011extended,cunningham2015linear}. For instance, Grassmann manifolds are well-explored in learning low-dimensional representation \citep{ngo2012scaled,dong2014clustering,zhang2018grassmannian}, and Stiefel manifolds are applied widely in fields like pattern recognition \citep{wright2009robust,browne2014orthogonal} and dimension reduction \citep{suzuki2010sufficient}. Other structures push forward low-rank approximations while guaranteeing high accuracy and computational feasibility \citep{vandereycken2010riemannian,sato2013riemannian,zhou2015rank}. 

Optimization on a Riemannian manifold has attracted great attention in the last decade, especially for large-scale learning tasks \citep{AbsilMahonySepulchre2008,bonnabel2013stochastic,sra2015conic}. Early research on stochastic gradient optimization on manifolds uses a fixed learning rate \citep{bonnabel2013stochastic,zhang2016riemannian,zhang2016first,ferreira2018gradient}. Some recent works have generalized the Euclidean-based adaptive learning schemes to the geometry-aware learning. \cite{RoyMhammediHarandi2018} proposed coordinate-wise adaptive operations with momentum based on \textsc{RMSProp}. \cite{becigneul2018riemannian} generalized the popular Euclidean adaptive learning rate schemes, \textsc{AdaGrad} and \textsc{Adam}, to adaptive learning on manifolds. \cite{kasai2019riemannian} adapted separate weight matrices corresponding to the row and column subspaces. 

It is a natural idea to treat special structures in variational parameters, such as symmetric positive definiteness for covariance matrix parameters or orthogonality constraints, as non-Euclidean spaces and apply Riemannian optimization. 
Existing methods usually deal with such constraints in an ad hoc manner that is specifically designed for the constraints under consideration \citep{Berg.et.al.2018}. It is worth mentioning that the natural gradient \citep{honkela2007natural,mishkin2018slang,zhang2018noisy,tran2019Variational} has been proposed as a speed-up gradient descent algorithm. The natural gradient, although operating on the Euclidean space, is a geometric object that is the steepest descent direction of the cost function while taking into account the geometry structure of the variational family. 

In this paper, we develop stochastic Gaussian variational approximation methods that guarantee the unique optimal variational parameters in the covariance matrix factorization. We do so by imposing constraints on the variational parameters and exploit their geometry to perform manifold-assisted optimization. More specifically, we factorize the covariance matrix using an LR+D structure, and ensure the uniqueness of that structure by introducing two manifold constraints. The main contributions of this work are:
\begin{itemize}
\item We combine the idea of Riemannian optimization with Euclidean adaptive stochastic gradients techniques, and propose two adaptive learning algorithms, namely \textsc{RGD-RMSProp} and \textsc{RGD-AdaDelta}, for stochastic optimization on manifolds. The two new optimizers enable computational speed-up in large-scale variational inference learning.
\item We impose the Stiefel and Grassmann manifold constraints on the parameters in the LR+D structure for the Gaussian variational approximation. The optimization on the Grassmannian equivalence classes solves the non-identifiability issue arising from the LR+D structure. It is possible to generalize the method to other constrained variational families such as the Sylvester normalizing flow of \cite{Berg.et.al.2018}.
\item We implement our methods on both low and high-dimensional cases to empirically test their performance. Our methods outperform baseline variational inference methods in terms of approximation accuracy and computational speed. 
\end{itemize}

The paper is organized as follows. Section~\ref{Sec:2} reviews the \textsc{VAFC} method, Stiefel and Grassmann manifolds, and Riemannian gradient descent. Section~\ref{Sec:3} details our Riemannian stochastic gradient schemes for Gaussian variational approximation. Section~\ref{Sec:4} empirically examines the performance of our methods. Section~\ref{Sec:5} concludes the paper with potential future extensions.

\section{Preliminary}
\label{Sec:2}
This section introduces the key concepts and sets up the notation. We start from Section~\ref{Sec:2.1} to formulate the objective function in variational inference. The \emph{low rank plus diagonal} (LR+D) factorization on the covariance matrix is discussed and its non-identifiability is identified. Two related manifolds, the \emph{Stiefel} and \emph{Grassmann} manifolds, are presented  in Section~\ref{Sec:2.2} and \ref{Sec:2.3}, respectively.  Section~\ref{Sec:2.4} describes the general Riemannian stochastic gradient decent method. 

\subsection{Gaussian Variational Approximation} \label{Sec:2.1}
Consider the Gaussian variational approximation problem for Bayesian inference.
Let $q_{\lambda}(\boldsymbol{\theta})$ denote a member of the variational family where $\lambda$ denotes the variational parameters. For example, if the variational family is multivariate Gaussians, then $\lambda$ consists of a mean vector and a covariance matrix. The objective is to find the best approximation $q_{\lambda^*}(\boldsymbol{\theta})$ to the  posterior $p(\boldsymbol{\theta}|\mathbf{x})$ where the optimal variational parameter $\lambda^*$ is found by minimizing the \emph{Kullback-Leibler divergence} between $q_{\lambda}(\boldsymbol{\theta})$ and $p(\boldsymbol{\theta}|\mathbf{x})$
\begin{align*}
\text{KL}(q_\lambda(\boldsymbol{\theta}) \| p(\boldsymbol{\theta} | \mathbf{x}))
&= \mathbb{E}[\log q_\lambda(\boldsymbol{\theta})]  - \mathbb{E}[\log p(\boldsymbol{\theta}, \mathbf{x})] + \log p(\mathbf{x}) = - \operatorname{ELBO}_q + \log p(\mathbf{x}).
\end{align*}
Due to the intractable evidence term $\log p(\mathbf{x})$, computing the KL-divergence is generally impossible. Fortunately, as this evidence term does not involve $q_{\lambda}(\boldsymbol{\theta})$, minimizing the KL-divergence is equivalent to maximizing the variational \emph{evidence lower bound} (ELBO). Formally, we rewrite this ELBO objective function as
\begin{align}
\mathcal{L}(\lambda) =& \int \log\frac{p(\boldsymbol{\theta})p(\mathbf{x}|\boldsymbol{\theta})}{q_{\lambda}(\boldsymbol{\theta})} q_{\lambda}(\boldsymbol{\theta})d\boldsymbol{\theta}  
=\; \mathbb{E}_q[\log h(\boldsymbol{\theta}) - \log q_{\lambda}(\boldsymbol{\theta})],
\label{Eq:Proj118-1}
\end{align}
where $h(\boldsymbol{\theta}) = p(\boldsymbol{\theta})p(\mathbf{x}|\boldsymbol{\theta})$ with the prior $p(\boldsymbol{\theta})$ and the model likelihood $p(\mathbf{x}|\boldsymbol{\theta})$. 
We denote by $m$ the dimension of $\lambda$, which can be considerably large. 
To find the optimal variational parameters from \eqref{Eq:Proj118-1}, we apply the stochastic gradient descent method by repeating
\[
\lambda^{(t+1)} = \lambda^{(t)} + \alpha_t \widehat{\nabla_{\lambda}\mathcal{L}}
\]
until the sequence converges. Here $\alpha_t$ is the learning step size at the $t$-th iteration, and $\widehat{\nabla_{\lambda}\mathcal{L}}$ is an unbiased approximation of the exact gradient $\nabla_{\lambda}\mathcal {L}\left(\lambda^{(t)}\right)$.

In the context of Gaussian variational approximation, the variational distribution to be estimated is $q_{\lambda}(\boldsymbol{\theta}) = \mathcal{N}(\boldsymbol{\mu}, \boldsymbol{\Sigma})$, $\lambda=( \boldsymbol{\mu},\boldsymbol{\Sigma})$. To preserve some off-diagonal covariance structure with a small computational cost, \cite{ong2018gaussian} factorize the full matrix using the LR+D representation, i.e.,
\begin{align}
\boldsymbol{\Sigma} = \mathbf{BB}^{\top} + \mathbf{D}^2_2,   \label{Eq:Sigma}
\end{align}
where $\mathbf{B}$ is an $m \times p$ full rank matrix with $p\leq m$ and $\mathbf{D}_2$ is a diagonal matrix with non-zero diagonal elements in vector $\mathbf{d}_2=(d_{21}, ..., d_{2m})^{\top}$. The reason for using the subscript in $\mathbf{D}_2$ will become clear shortly. We use the number of factors $p$ to control the complexity of $q_{\lambda}(\boldsymbol{\theta})$. With a small $p$ value, both the storage and computational cost are reduced on approximating the dense covariance matrix $\boldsymbol{\Sigma}$, so that it is possible to solve large-scale problems. Specifically, when $p=0$, the model degenerates to the mean-field variational structure. With the factorized form \eqref{Eq:Sigma}, the variational parameter $\lambda$ collects $\{\boldsymbol{\mu}, \mathbf B, \mathbf d_2\}$.
We rewrite \eqref{Eq:Proj118-1} as
\[
\mathcal{L}(\lambda) = \mathbb{E}_q[\log h(\boldsymbol{\theta})]+\frac12\log|\mathbf{BB}^{\top} + \mathbf{D}^2_2|+\frac{m}2(\log(2\pi) +1).
\]
We ignore the constant term $\frac{m}2(\log(2\pi) +1)$ since it is independent from $\lambda$, the parameters to be estimated. The final objective function is
\begin{align}
\mathcal{L}(\lambda) = \mathbb{E}_q[\log h(\boldsymbol{\theta})] + \frac12\log|\mathbf{BB}^{\top} + \mathbf{D}^2_2|. \label{Eq:New1}
\end{align}

To obtain efficient stochastic gradient estimates, we use the \emph{reparameterization trick} \citep{KingmaWelling2014}. Let $\mathbf z \in\mathbb{R}^p$ and $\boldsymbol{\epsilon}\in\mathbb{R}^m$ be two standard Gaussian variables, i.e., $(\mathbf z, \boldsymbol{\epsilon}) \sim f(\mathbf z, \boldsymbol{\epsilon}) = \mathcal{N}(\mathbf 0, \mathbb{I})$. The random variable $\boldsymbol{\theta} \sim \mathcal{N}(\boldsymbol{\mu}, \boldsymbol{\Sigma})$ can be reparameterized as
$$
\boldsymbol{\theta} = \boldsymbol{\mu} + \mathbf{Bz} + \mathbf{d}_2 \circ \boldsymbol{\epsilon} = \boldsymbol{\mu} + \mathbf{Bz} + \mathbf{D}_2 \boldsymbol{\epsilon},
$$ 
where $\circ$ indicates the Hadamard product. The objective function \eqref{Eq:New1} then becomes
\begin{align}
\mathcal{L}(\lambda) = \mathbb{E}_f[\log h(\boldsymbol{\mu} + \mathbf{Bz} + \mathbf{d}_2 \circ \boldsymbol{\epsilon})] + \frac12\log|\mathbf{BB}^{\top} + \mathbf{D}^2_2|.
\label{Eq:New2}
\end{align}

The advantage of \eqref{Eq:New2} over \eqref{Eq:New1} is that, when the expectation is approximated by a sampling method, the samples $(\mathbf z,\boldsymbol{\epsilon})$ are independent of $\lambda$, which can be explained as a control variate scheme \citep{paisley2012variational,pmlr-v89-xu19a}. 

Such LR+D factorization for Gaussian variational approximation has many attractive properties and enjoys competitive performance \citep{ong2018gaussian}. However, this VAFC method of \cite{ong2018gaussian} suffers from the non-identifiability issue.
Consider any orthogonal matrix $\mathbf Q$ of size $p\times p$, it is easy to see that
\begin{align}
|\mathbf{B}\mathbf{Q} (\mathbf{B}\mathbf{Q})^{\top}+\mathbf{D}^2_2| &=  |\mathbf{B}\mathbf{B}^{\top}+\mathbf{D}^2_2|, \notag\\ 
\mathbb{E}_f[\log h(\boldsymbol{\mu}+\mathbf{B(Qz)}+\mathbf{d}_2 \circ \boldsymbol{\epsilon})] &=
\mathbb{E}_f[\log h(\boldsymbol{\mu} + \mathbf{B z} + \mathbf{d}_2 \circ \boldsymbol{\epsilon})], \notag
\end{align}
as $\mathbf{Q}\mathbf{z}$ still follows the standard Gaussian distribution. Then for any orthogonal matrix $\mathbf Q$,
\begin{align}
\mathcal{L}(\lambda(\boldsymbol{\mu}, \mathbf B, \mathbf D_2))
=\mathcal{L}(\lambda(\boldsymbol{\mu}, \mathbf B\mathbf Q, \mathbf D_2)). \label{Eq:New3}
\end{align}
Hence, if $\lambda(\boldsymbol{\mu}, \mathbf B, \mathbf D_2)$ is the variational solution, so is $\lambda(\boldsymbol{\mu}, \mathbf B\mathbf Q, \mathbf D_2)$.
This non-identifiability may bring in certain optimization difficulties. While there are potential practical advantages that determining a small number of factors $p$ helps reduce the probability of occurring the identification problem, the method, from a purely statistical point of view, still dissatisfies the non-ignorable arbitrariness \citep{bartholomew2011latent}.

To avoid the potential non-identifiability issue, restrictions on the factor matrix $\mathbf{B}$ are often imposed. For example, $\mathbf{B}$ can be a column full-rank lower triangular matrix with positive diagonal elements \citep{ong2018gaussian,TranNguyenNottKohn2018}. Unfortunately, without careful treatment, the updating process might fail to guarantee such constraints on $\mathbf{B}$.
Consider a stochastic gradient process with a fixed learning rate $\alpha$ = 0.01. Suppose we have $\mathbf{B}^{(t)}$ = 
$\begin{bmatrix}
1 & 0 \\
2 & 1 
\end{bmatrix}$ and a calculated gradient $\frac{\partial \mathcal{L}(\mathbf{B})}{\partial \mathbf{B}}$ = 
$\begin{bmatrix}
-2 & 0 \\
-50 & 100 
\end{bmatrix}$. This gives the updated $\mathbf{B}^{(t+1)} = 
\begin{bmatrix}
1.02 & 0 \\
2.5 & 0 
\end{bmatrix}$, which is a non-feasible solution to the optimization problem because the $\mathbf{B}^{(t+1)}$ is rank-deficient.

To this end, we propose an optimization framework on manifolds to guarantee a feasible solution that respects the necessary restrictions. We do so by utilizing the concepts of Stiefel manifold and Grassmann manifold that we review next.

\begin{rem}\label{Remark1} 
We do not consider the structure in \cite{journee2010low} as their manifold is not complete. The geometry is only suggested for Fixed-rank situations, where the rank of $\mathbf{B B}^{\top}$ equals exactly to some $k$ ($k \leq m$). The method also failed in our empirical experiments. Evidence is provided in the first row of Figure~\ref{fig-1a} where the algorithm does not converge properly. 
\end{rem}

\subsection{Stiefel Manifold Constraint}
\label{Sec:2.2}
\begin{definition}[The Stiefel Manifold \citep{AbsilMahonySepulchre2008}] Let $p\leq m$, the Stiefel manifold $\mathcal{S}(p,m)$ is the set of $m\times p$-dimensional matrices consisting of orthonormal columns. That is
$$
\mathcal{S}(p,m) = \{\mathbf{B}: \;\; \mathbf{B}^{\top}\mathbf{B}=\mathbb{I}_p\}.
$$
\end{definition}

A Stiefel manifold $\mathcal{S}(p,m)$ consists of all $p$ orthogonormal basis vectors that are rigidly connected to each other and represents all the equivalent results, i.e., a Stiefel manifold point $\mathbf{B} \in \mathcal{S}(p,m)$ includes all the equivalent $\mathbf{BQ}$ by a set of orthonormal bases. Formally, the covariance matrix in the Gaussian variational approximation  $q_{\lambda}(\boldsymbol{\theta}) = \mathcal{N}(\boldsymbol{\mu}, \boldsymbol{\Sigma})$ is factorized as
\begin{align}
\boldsymbol{\Sigma} = \mathbf{B D^2_1 B}^{\top} + \mathbf{D}^2_2, \label{Eq:Proj118-2}
\end{align} 
where $\mathbf{D}_1$ and $\mathbf{D}_2$ are diagonal matrices of size $p\times p$ and $m\times m$. We introduce $\mathbf{D}_1$ to solve the non-identifiability issue for the parameter $\mathbf B$. For the new covariance in \eqref{Eq:Proj118-2}, its corresponding Gaussian $q_{\lambda}(\boldsymbol{\theta}) = \mathcal{N}(\boldsymbol{\mu}, \boldsymbol{\Sigma})$ can be found by a new deterministic transformation
\begin{align}
\boldsymbol{\theta} = \boldsymbol{\mu} + \mathbf{BD_1 z} + \mathbf{d}_2 \circ \boldsymbol{\epsilon}
\label{Eq:Proj118-2a}
\end{align}
where $(\mathbf{z}, \boldsymbol{\epsilon})$ follows the standard $(p+m)$-dimensional Gaussian distribution $f(\mathbf{z}, \boldsymbol{\epsilon})$.

\begin{rem} 
We adopt the scaling factor $\mathbf{D}_1$ in \eqref{Eq:Proj118-2} to compensate for the loss of scale from the orthogonality requirement. Also, the factor effectively prevents the non-identifiability of $\mathbf{B}$.
\end{rem}

Applying the reparameterization trick in \eqref{Eq:Proj118-2a} will update the objective function \eqref{Eq:New1} to
\begin{align}
\mathcal{L}(\lambda) =& \mathbb{E}_f\left[\log h(\boldsymbol{\mu} + \mathbf{BD_1 z} + \mathbf{d}_2 \circ \boldsymbol{\epsilon})\right] 
+\frac12\log\left|\mathbf{B D^2_1 B}^{\top} + \mathbf{D}^2_2\right|,
\label{Eq:Proj118-3} 
\end{align} 
where the expectation $\mathbb{E}_f$ is with respect to the standard Gaussian distribution $f(\mathbf{z}, \boldsymbol{\epsilon})=\mathcal{N}(0, \mathbb{I})$.

The Stiefel manifold constraint guarantees a full column rank solution with an orthonormal $\mathbf{B}$. The stochastic gradient estimates on $\mathbf{B}$ will be conducted with its manifold geometry.

\subsection{Grassmann Manifold Constraint}
\label{Sec:2.3}
\begin{definition}[The Grassmann Manifold \citep{AbsilMahonySepulchre2008}] The Grassmann manifold, denoted by $\mathcal{G}(p,m)$, consists of all the $p$-dimensional subspaces in $\mathbb{R}^m$ ($p\leq m$). Any Grassmann point on $\mathcal{G}(p,m)$ can be represented by an $m\times p$ matrix $\mathbf{B}$ with orthonormal columns, that is, $\mathbf{B}^{\top}\mathbf{B}=\mathbb{I}_p$.
\end{definition}

A Grassmann point on $\mathcal{G}(p,m)$ is the subspace spanned by the columns of the factor matrix $\mathbf{B}$. It can be considered as an equivalence class over Stiefel manifold $\mathcal{S}(p,m)$, i.e., for a representative $\mathbf{B}\in \mathcal{S}(p,m)$, it has an equivalent class
\[
[\mathbf{B}] = \{\mathbf{BQ}:\;\; \mathbf{B}\in \mathcal{S}(p,m)  \text{ and } \mathbf{Q}\in\mathcal{O}(p)\},
\]
where $\mathcal{O}(p)$ is the orthogonal group of order $p$ \citep{AbsilMahonySepulchre2008}. 

The property in \eqref{Eq:New3} shows that the ELBO $\mathcal{L}(\lambda)$ in \eqref{Eq:New2} is a well defined function on $\mathbb{R}^m \otimes \mathcal{G}(p,m)\otimes \mathbb{R}^m$ where $\boldsymbol{\mu}\in\mathbb{R}^m$, $\mathbf B\in\mathcal{G}(p,m)$ and $\mathbf d_2\in\mathbb{R}^m$. We then rewrite \eqref{Eq:New1} as
\begin{align}
\mathcal{L}(\lambda) = & \mathbb{E}_f\left[\log h(\boldsymbol{\mu} + \mathbf{Bz} + \mathbf{d} \circ \boldsymbol{\epsilon})\right] + \frac12\log\left|\mathbf{BB}^{\top}+\mathbf{D}^2\right|. \label{Eq:Proj118-3a}
\end{align} 
The equation is arranged similarly to \eqref{Eq:New2}, except $\mathbf{B}$ is a Grassmann point, i.e., a representative Stiefel matrix of its equivalence class. The optimal solution $\mathbf{B^*}$ from \eqref{Eq:Proj118-3a} is a Grassmann point with all equivalence solutions $[\mathbf{B^*}]$, and it is free from the identification problem. In contrast, \eqref{Eq:New2} searches in the Euclidean space for a single solution $\mathbf{B}$ (among many) instead of its unique equivalence class solution $[\mathbf{B}]$. 

Due to the manifold constraints, the conventional gradient update rules working on the Euclidean space fail to guarantee convergence for the objective functions in \eqref{Eq:Proj118-3} and \eqref{Eq:Proj118-3a}. Instead, we will apply the Riemannian gradient descent rules that we review next.

\subsection{Riemannian Gradient Descent}
\label{Sec:2.4}
We have formulated two new objective functions with the Stiefel and the Grassmann manifold constraints respectively. This section reviews how an optimization problem on manifolds is solved by standard Riemannian optimization algorithms. We start from the basic line-search method that involves a one-step gradient descent. The notion of \emph{retraction mapping} will be introduced to enable updating on a manifold. We then review the gradient momentum methods that help accelerate convergence. By \emph{vector transport}, we accumulate tangent vectors from different vector spaces onto the same space. 
Further, we show how a point on a manifold is connected to a tangent space by \emph{projection}.

\begin{figure}[tbh]
\centering
\includegraphics[width = 0.75\textwidth, trim = {4cm 16.7cm 3cm 5.7cm}, clip]{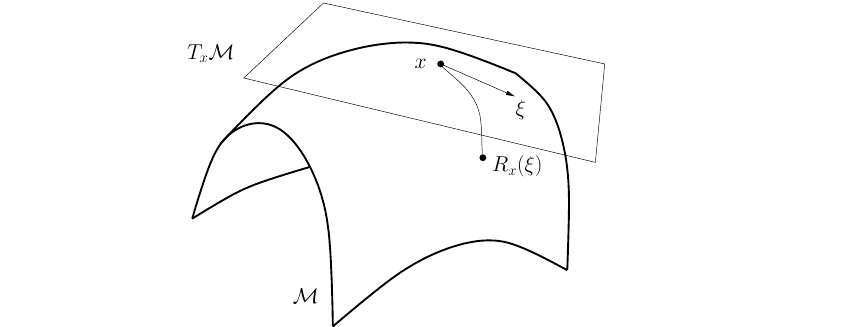}
\caption{Illustration of the retraction operation on the tangent space. $\mathcal{M}$ is the manifold. $x$ is the points on the manifold. $\xi$ is the tangent vector on the tangent space $T_x\mathcal{M}$, and $\mathrm{R}_x(\xi)$ is the retraction of $\xi$ back to the manifold $\mathcal{M}$ \citep{AbsilMahonySepulchre2008}.} 
\label{retraction}
\end{figure}

For a differentiable function $\mathcal{L}(x)$ in an Euclidean space, the line-search method defines
$$
x_{k+1} = x_{k}+\alpha_{k} \xi_{k},
$$
where $\xi_{k} \in \mathbb{R}^{m}$ is the search direction and $\alpha_{k} \in \mathbb{R}_+$ is the step size. The gradient, however, relies on the vector space structure of $\mathbb{R}^{m}$, and this dependency is not fundamental for an objective function defined on nonlinear manifolds. Instead, for objective functions defined on a manifold $\mathcal{M}$, the general update rule is
$$
x_{k+1} = \mathrm{R}_{x_k}(\alpha_{k} \xi_{k}),
$$
where $\xi_{k}$ is a tangent vector to $\mathcal{M}$ at $x_{k}$. We define $\mathrm{R}_{x}$ as a retraction at ${x}$ to map $\xi_{k}$ smoothly from the tangent space $T_x\mathcal{M}$ onto $\mathcal{M}$. The retraction mapping guarantees $x$ is continuously updated along the steepest descent \citep{AbsilMahonySepulchre2008}, i.e., its moving direction is on a tangent vector, while the point is staying on the manifold. Figure.~\ref{retraction} visualizes the process of a retraction mapping to preserve the gradient at $x$.

In the line-search method, the tangent vector $\xi$ is the steepest-descent direction of the objective function, i.e., $\xi_k = -\text{grad}\mathcal{L}(x_k)$ the Riemannian gradient of $\mathcal{L}$ at $x_k$. Many modern gradient descent methods include momentum to dampen oscillations by accumulating the stochastic gradient descents. In Euclidean space, the search direction is produced by summing up $-\text{grad}\mathcal{L}(x_k)$ and $\xi_{k-1}$. On a nonlinear manifold, the two tangent vectors with respect to $x_k$ and $x_{k-1}$ do not belong to the same subspace, thus they cannot be combined arbitrarily. This issue can be fixed by applying a vector transport to bring $\xi_{k-1}$ to the tangent space at $x_k$. The search direction is then
$$
\xi_k = -\text{grad}\mathcal{L}(x_k) + \alpha_k \Gamma_{x_{k-1}\rightarrow x_k}(\xi_{k-1}),
$$
where $\Gamma_{x_{k-1}\rightarrow x_k}$ is the parallel transport that maps a vector $\xi_{k-1} \in T_{x_{k-1}}\mathcal{M}$ to a vector  in $T_{x_{k}}\mathcal{M}$.

For manifolds that can be embedded in the ambient Euclidean space, their Riemannian gradient is found by projecting the Euclidean gradient in the ambient space of the manifold onto the tangent space at the point $x$. The way to implement the projection operator depends on manifolds. Table~\ref{table-1} and Table~\ref{table-2} summarize the mapping formulas of Stiefel and Grassmann manifolds. 

\begin{table*}[tbh]
\begin{center}
\begin{tabular}{ll}
Manifolds & Stiefel $\mathcal{S}(m,p)$ \\ \hline
Tangent Spaces & $T_B\mathcal{S}(m,p)=\{U\in\mathbb{R}^{m\times p}: \text{sym}(B^{\top}U)=0\}$ \\ 
Projection $\pi_B(Z)$ & $Z-B\text{sym}(B^{\top}Z) $ \\ 
Retraction $\mathrm{R}_B(U)$ & $(B+U)(I+U^{\top}U)^{-\frac12}$ \\ 
Parrallel Transport $\Gamma_{B_1\rightarrow B_2}(U)$ & $\pi_{B_2}(U)$ \\ 
\end{tabular}
\vspace{12pt}
\caption{Riemannian operations for the Stiefel manifold structure.}
\label{table-1}
\end{center} 
\end{table*}

\begin{table*}[tbh]
\begin{center}
\begin{tabular}{llll}
Manifolds & Grassmann $\mathcal{G}(m,p)$ &&\\ \hline
Tangent Spaces & $T_B\mathcal{G}(m,p) = \{U\in\mathbb{R}^{m\times p}:,B^{\top}U = 0\}$ &&\\ 
Projection $\pi_B(Z)$ & $(I - BB^{\top})Z$ &&\\ 
Retraction $\mathrm{R}_B(U)$ & $\text{polar}(B+U)$ &&\\ 
Parrallel Transport $\Gamma_{B_1\rightarrow B_2}(U)$ & $\pi_{B_2}(U)$ &&\\ 
\end{tabular}
\vspace{12pt}
\caption{Riemannian operations for the Grassmann manifold structure.}
\label{table-2}
\end{center} 
\end{table*}

Note that we define the symmetric operation of $\mathbf{B}^{\top}\mathbf{U}$ as $\text{sym}(\mathbf{B}^{\top}\mathbf{U}) = \frac{1}{2}(\mathbf{B}^{\top}\mathbf{U} + \mathbf{U}^{\top}\mathbf{B})$. The retraction $\mathrm{R}_\mathbf{B}(\mathbf{U})$ on the Grassmann manifold can be implemented as the polarization $\text{polar}(\mathbf{B}+\mathbf{U})$. When $\mathbf{U} \neq 0$, the matrix $\mathbf{B}+\mathbf{U}$ is not orthogonal. To retain a representative on Grassmann manifold, we use the polar decomposition to avoid ill-conditioning.

In a nutshell, the Riemannian gradient descent algorithm is applied to update a point $x$ on a manifold. The search direction is found by the Riemannian gradient over the objective function, which projects the Euclidean gradient onto the tangent space. For vectors from different tangent spaces, parallel transport is adopted to accumulate them. Finally, the searching variable is mapped back to the manifold space by the retraction operator.

\section{Approximation with Riemannian Operations}\label{Sec:3}
We have reformulated the objective functions in \eqref{Eq:Proj118-3} and \eqref{Eq:Proj118-3a} in Section~\ref{Sec:2.2} and \ref{Sec:2.3}, respectively. The task is to find the optimal variational parameters $\lambda^*=\{\boldsymbol{\mu}^*, \mathbf{B}^*, \mathbf{D}^*_1, \mathbf{D}^*_2\}$ that maximize the objective function $\mathcal{L}(\lambda)$. For the unconstrained parameters $\{\boldsymbol{\mu}, \mathbf{D}_1, \mathbf{D}_2\}$, we apply the \emph{stochastic gradient descent} (SGD) update rules in the Euclidean space, see Section~\ref{Sec:3.1}. In Section~\ref{Sec:3.2}, we introduce the \emph{Riemannian (stochastic) gradient descent} (RGD) update rules for the constrained parameter $\mathbf{B}$. We consider four different variants, where the first two are conventional methods for Riemannian optimization, and the last two adaptive update rules are first developed in this paper. 

\subsection{Euclidean Gradient of the Log-Likelihood Function}
\label{Sec:3.1}
We first calculate the Euclidean partial derivative of the objective function with respect to each parameter. The unconstrained parameters can be updated directly with the SGD rules, while the constrained parameters require these derivatives as the input to implement the RGD rules. In the case of Stiefel constraint in \eqref{Eq:Proj118-3}, we calculate the derivatives as follow
\begin{align} 
&\hspace{-0.6cm}\nabla_{\boldsymbol{\mu}} \mathcal{L} = \mathbb{E}_f[\nabla_{\boldsymbol{\theta}} \log h(\boldsymbol{\mu} + \mathbf{BD_1 z} + \mathbf{d}_2 \circ \boldsymbol{\epsilon})]; 
\label{Eq:Stiefel1} \\
&\hspace{-0.6cm}\nabla_{\mathbf{B}} \mathcal{L}  = \mathbb{E}_f[\nabla_{\boldsymbol{\theta}} \log h(\boldsymbol{\mu} + \mathbf{BD_1 z} + \mathbf{d}_2 \circ \boldsymbol{\epsilon}) (\mathbf{z}\circ \mathbf{d}_1)^{\top}] + (\mathbf{B D^2_1 B}^{\top} + \mathbf{D}^2_2)^{-1} \mathbf{B D}^2_1;
\label{Eq:Stiefel2} \\
&\hspace{-0.6cm}\nabla_{\mathbf{d}_1} \mathcal{L}  = \mathbb{E}_f[(B^{\top}\nabla_{\boldsymbol{\theta}} \log h(\boldsymbol{\mu} + \mathbf{BD_1 z} + \mathbf{d}_2 \circ \boldsymbol{\epsilon})) \circ \mathbf{z}] + \text{diag}(B^{\top}(\mathbf{B D^2_1 B}^{\top} + \mathbf{D}^2_2)^{-1}\mathbf{B}) \circ \mathbf{d}_1;
\label{Eq:Stiefel3}\\
&\hspace{-0.6cm}\nabla_{\mathbf{d}_2} \mathcal{L}  = \mathbb{E}_f[\text{diag}(\nabla_{\boldsymbol{\theta}} \log h(\boldsymbol{\mu} + \mathbf{BD_1 z} + \mathbf{d}_2 \circ \boldsymbol{\epsilon}) \boldsymbol{\epsilon}^{\top})] + \text{diag}((\mathbf{B D^2_1 B}^{\top} + \mathbf{D}^2_2)^{-1})\circ \mathbf{d}_2,
\label{Eq:Stiefel4}
\end{align}
where $\nabla_{\boldsymbol{\theta}} \log h(\boldsymbol{\mu} + \mathbf{BD_1 z} + \mathbf{d}_2 \circ \boldsymbol{\epsilon}) = \nabla_{\boldsymbol{\theta}} \log h(\boldsymbol{\theta})$, which is easy to calculate, e.g., by automatic differentiation. For high-dimensional problems, it is challenging to compute the inverse matrix $(\mathbf{B D^2_1 B}^{\top} + \mathbf{D}^2_2)^{-1}$ efficiently. Instead, we follow \cite{ong2018gaussian} and apply the \emph{Woodbury identity} \citep{woodbury1950inverting} to  convert the inverse of an $m\times m$ matrix to an inverse of a $p\times p$ matrix as follows
\begin{align*} 
&(\mathbf{B D^2_1 B}^{\top} + \mathbf{D}^2_2)^{-1} = \mathbf{D}^{-2}_2 - \mathbf{D}^{-2}_2 \mathbf{BD}_1(\mathbb{I}_p+\mathbf{D_1B^{\top}D}^{-2}_2\mathbf{BD}_1)^{-1}\mathbf{D_1B^{\top}D}^{-2}_2.
\end{align*}
Similarly, for \eqref{Eq:Proj118-3a} with the Grassmann manifold constraint, we have 
\begin{align}
\nabla_{\boldsymbol{\mu}} \mathcal{L}  &= \mathbb{E}_f[\nabla_{\theta} \log h(\boldsymbol{\mu} + \mathbf{B z} + \mathbf{d}_2 \circ \boldsymbol{\epsilon})]; 
\label{Eq:Grass1}\\
\nabla_{\mathbf{B}} \mathcal{L}  &= \mathbb{E}_f[\nabla_{\theta} \log h(\boldsymbol{\mu} + \mathbf{B z} + \mathbf{d}_2 \circ \boldsymbol{\epsilon}) \mathbf{z}^{\top}] + (\mathbf{B B}^{\top} + \mathbf{D}^2_2)^{-1} \mathbf{B};
\label{Eq:Grass2}\\
\nabla_{\mathbf{d}_2} \mathcal{L}  &= \mathbb{E}_f[\nabla_{\theta} \log h(\boldsymbol{\mu} + \mathbf{B z} + \mathbf{d}_2 \circ \boldsymbol{\epsilon}) \boldsymbol{\epsilon}^{\top}] + \text{diag}((\mathbf{B B}^{\top} + \mathbf{D}^2_2))^{-1})\circ \mathbf{d}_2.
\label{Eq:Grass4}
\end{align}

\subsection{Riemannian Stochastic Gradient Descent}
\label{Sec:3.2}
We now describe four gradient-based updating rules under the RGD framework. We start with the vanilla RGD method \citep{bonnabel2013stochastic} that uses a fixed learning rate. To accelerate convergence, we consider the second RGD update rule with momentum \citep{RoyHarandi2017}. 
We propose two adaptive learning rules for RGD, constrained \textsc{RMSProp} rule and the simulated \textsc{AdaDelta} rule. Both methods use a new learning rate for every parameter at every time step, and this learning rate is determined automatically during the update process. These two adaptive learning rules for Riemannian stochastic optimization, to the best of our knowledge, are first developed in this paper.
In the following, we will use the notation $\mathcal{L}(\mathbf B)$ for $\mathcal{L}(\lambda)$ where we consider the optimization over the manifold variable $\mathbf B$.

\subsubsection{Basic RGD method with Fixed Learning Rate}
The first update scheme follows the procedure in \cite{bonnabel2013stochastic} and extends the vanilla stochastic gradient descent algorithm to a Riemannian manifold. As reviewed in Section~\ref{Sec:2.4}, both Stiefel and Grassmann manifolds can be embedded in the ambient Euclidean space. The Riemannian gradient of the objective function at a point $\mathbf{B}$ is calculated by projecting its Euclidean gradient $\nabla_{\mathbf B} \mathcal{L}(\mathbf{B})$ from the ambient space to the tangent space, i.e.,
\begin{align}
\text{grad}\mathcal{L}(\mathbf{B}) = \pi_\mathbf{B}(\nabla_{\mathbf B} \mathcal{L}(\mathbf{B})). \label{Eq:Proj118-6}
\end{align}
The explicit projection mapping function $\pi_\mathbf{B}$ of the two manifold constraints are defined in Table~\ref{table-1} and Table~\ref{table-2}. Once the Riemannian gradient is prepared, the RGD update rule is applied
\begin{align}
\mathbf{B}^{(t+1)} = \mathrm{R}_{\mathbf{B}^{(t)}}\left(-\alpha \left.\text{grad}\mathcal{L}(\mathbf{B})\right|_{\mathbf{B}^{(t)}}\right), \label{Eq:RGD}
\end{align}
where $\mathrm{R}_\mathbf{B}(\cdot)$ is the retraction operator that maps the point on tangent space onto the manifold, and $\alpha$ is a fixed learning rate.
We refer to this basic RGD method as  \textsc{RGD-Basic}.

\subsubsection{The Constrained SGD with Momentum Rule}
The basic RGD updating rule in \eqref{Eq:RGD} provides an intuitive way of optimization over manifolds. However, that line-search method over the steepest-descent direction could be inefficient around local optima where the cost surface is usually steeper in one dimension than in another. Based on the momentum idea from the SGD literature, , \cite{RoyHarandi2017} proposed Constrained RGD with Momentum (\textsc{cRGD-M}). 
Using their method, the manifold-constrained variable $\mathbf{B}$ is updated by
\begin{align}
\mathbf{m}^{(t+1)} =& \beta \Gamma_{\mathbf{B}^{(t-1)}\rightarrow \mathbf{B}^{(t)}}(\mathbf{m}^{(t)}) 
+\alpha \left.\text{grad}\mathcal{L}(\mathbf{B})\right|_{\mathbf{B}^{(t)}} \label{Eq:M} \\
\mathbf{B}^{(t+1)} =&  \mathrm{R}_{\mathbf{B}^{(t)}}(-\mathbf{m}^{(t+1)})
\label{Eq:cRGD-M}
\end{align}
where $0<\beta$ is the momentum constant and $\Gamma_{\mathbf{B}_1\rightarrow \mathbf{B}_2}(\mathbf{U})$ is the vector transport which maps the tangent vector $\mathbf{U}$ from the tangent space at $\mathbf{B}_1$ to that at $\mathbf{B}_2$. The implementation of this vector transport can be found in Table~\ref{table-1} and Table~\ref{table-2}.

\subsubsection{The Modified Constrained \textsc{RMSProp} Rule}
Tuning a proper learning rate can be tricky especially for large-scale learning tasks. Also, the same learning rate might not apply to all parameter updates during the training process. \cite{RoyMhammediHarandi2018} considered the parameter-wise adaptive learning rate for the sparse data structure so that the optimizer gains a rapid convergence speed. Their method is a constrained \textsc{RMSProp} rule that ensures $\mathbf{B}^{\top}\mathbf{B} = \mathbb{I}$. However, this update rule breaks when the combination of a parallel transport and a projection on the tangent space becomes negative. Instead, we modify the update rule in \cite{RoyMhammediHarandi2018} and propose \textsc{RGD-RMSProp} as follows. We include the absolute operator over the element $E(g_\mathbf{B}^2)$ and compensate its sign after the square root operation. 
\begin{align}
&\mathbf{B}^{(t+1)} 
= \mathrm{R}_{\mathbf{B}^{(t)}}\left(-\alpha\pi_{\mathbf{B}^{(t)}}\left(\frac{\nabla\mathcal{L}(\mathbf{B}^{(t)})}{\text{sgn}(E(g^2_\mathbf{B})^{(t+1)})\odot\sqrt{|E(g^2_\mathbf{B})^{(t+1)}|}+\epsilon}\right)\right)    \label{Eq:Proj118-7}
\end{align}
where
\begin{align} 
 E(g^2_\mathbf{B})^{(t+1)} =& \beta \Gamma_{\mathbf{B}^{(t-1)}\rightarrow \mathbf{B}^{(t)}} \left(E(g^2_\mathbf{B})^{(t)}\right) 
 + (1-\beta) \pi_{\mathbf{B}^{(t)}}\left(\nabla \mathcal{L}(\mathbf{B}^{(t)}) \odot \nabla \mathcal{L}(\mathbf{B}^{(t)}) \right) \label{Eq:G2}
\end{align}
with $\odot$ as the element-wise product of two matrices and $\nabla \mathcal{L}$ as the Euclidean gradient of the objective function. For the empirical experiments, we suggest $\alpha=0.05, \beta = 0.95$ and $\epsilon=10^{-6}$.

\begin{rem} Our \textsc{RGD-RMSProp} can be regarded as the basic RGD with a constrained normalized Euclidean gradient. The method differs from the Euclidean \textsc{RMSProp} \citep{HintonSrivastavaSwersky2016} and requires sophisticated mappings to accelerate multiple tangent vectors.
\end{rem}

\subsubsection{The Simulated \textsc{AdaDelta} Rule}
Compared to the basic RGD, the \textsc{RGD-RMSProp} includes the adaptive learning rate scheme that effectively speeds up convergence. However, \eqref{Eq:Proj118-7} has two limitations. First, selecting the learning rate $\alpha$ can be challenging. Second, when updating the parameter $\mathbf B$, the hypothetical units are related to the gradients $\nabla \mathcal{L}(\mathbf B)$ rather than the parameters \citep{Zeiler2012}. To fix the two issues, we propose using element-wise division in \eqref{Eq:G2}. The update rule is arranged as follows
\begin{align}
 E(g^2_\mathbf{B})^{(t)} =& \beta \Gamma_{\mathbf{B}^{(t-1)}\rightarrow \mathbf{B}^{(t)}} \left(E(g^2_\mathbf{B})^{(t-1)}\right) 
 + (1-\beta) \pi_{\mathbf{B}^{(t)}}\left(\nabla \mathcal{L}(\mathbf{B}^{(t)}) \odot \nabla \mathcal{L}(\mathbf{B}^{(t)}) \right); &\label{Eq:B3}\\
\Delta \mathbf{B}^{(t)} =& \frac{\text{sgn}(E(\Delta \mathbf{B}^2)^{(t-1)})\odot\sqrt{|E(\Delta \mathbf{B}^2)^{(t-1)}|} + \epsilon}{\text{sgn}(E(g^2_{\mathbf B})^{(t)})\odot\sqrt{|E(g^2_{\mathbf{B}})^{(t)}|}+\epsilon} 
\odot \nabla \mathcal{L}(\mathbf{B}^{(t)});   &\label{Eq:B1} \\
E(\Delta \mathbf{B}^2)^{(t)} =& \beta \Gamma_{\mathbf{B}^{(t-2)}\rightarrow \mathbf{B}^{(t-1)}} \left(E(\Delta \mathbf{B}^2)^{(t-1)}\right) 
+(1-\beta) \pi_{\mathbf{B}^{(t-1)}}\left(\Delta \mathbf{B}^{(t)}  \odot \Delta \mathbf{B}^{(t)}  \right); &\label{Eq:B2}\\
\mathbf{B}^{(t+1)} =& \mathrm{R}_{\mathbf{B}^{(t)}}\left(-\pi_{\mathbf{B}^{(t)}}(\Delta \mathbf{B}^{(t)})\right). \label{Eq:Proj118-9} 
\end{align}
At each step, the learning rate is self-updated with the previous performance of the gradient. The update rules above could be considered as the Riemannian version of its Euclidean counterpart \citep{Zeiler2012}.
We refer to this method as 
\textsc{RGD-AdaDelta}.

The overall algorithm for the Manifold assisted Gaussian Variational Approximation algorithm is summarized in Algorithm \ref{Alg:1}.
For notational simplicity, in the following, we use S for Stiefel and G for Grassmann.

\begin{algorithm}[H]
\caption{Manifold Assisted Gaussian Variational Approximation Algorithm} \label{Alg:1}
\begin{algorithmic}[1] 
\REQUIRE Initialize $\lambda \leftarrow (\boldsymbol{\mu}^{(0)}, \mathbf{B}^{(0)}, \mathbf{d}_1^{(0)}, \mathbf{d}_2^{(0)}),\ t \leftarrow 0.$
\ENSURE $\lambda^* =\{\boldsymbol{\mu}^*, \mathbf{B}^{*}, \mathbf{d}_1^*, \mathbf{d}_2^{*}\}$ where $\mathbf{B}^*$ is a Stiefel or Grassmann Point, i.e., $\mathbf{B}^{*T}\mathbf{B}^*=\mathbb{I}_p$.
\IF{not stopping}

\STATE Sample $(z^{(t)}, \epsilon^{(t)}) \sim N(0, I)$;
\STATE Estimate  the  Euclidean gradients  at $\lambda^{(t)}=\{\boldsymbol{\mu}^{(t)}, \mathbf{B}^{(t)}, \mathbf{d}_1^{(t)}, \mathbf{d}_2^{(t)}\}$ by: \\
\eqref{Eq:Stiefel1} - \eqref{Eq:Stiefel4} for the Stiefel Constraint, OR\\
\eqref{Eq:Grass1} - \eqref{Eq:Grass4} for the Grassmann Constraint;
\STATE Update the non-constrained \{$\boldsymbol{\mu}^{(t+1)}, \mathbf{d}_1^{(t+1)}, \mathbf{d}_2^{(t+1)}$\} with the Euclidean SGD rules;
\STATE Calculate the Riemann gradient with respect to constrained parameter $\mathbf{B}$ by \eqref{Eq:Proj118-6};
\STATE Prepare auxilliary variables by:\\
\eqref{Eq:M} (\textsc{cRGD-M}) for $m^{(t+1)}$, OR \\
\eqref{Eq:G2} (\textsc{RGD-RMSProp}) for $E(g^2_\mathbf{B})^{(t+1)}$, OR \\
\eqref{Eq:B3} - \eqref{Eq:B2} (\textsc{RGD-AdaDelta}) for $E(g^2_\mathbf{B})^{(t)}$, $\Delta \mathbf{B}^{(t)}$ and $E(\Delta \mathbf{B}^2)^{(t)}$;
\STATE Update $\mathbf{B}^{(t+1)}$ by: \\
\eqref{Eq:RGD} (\textsc{RGD-Basic}), OR \\
\eqref{Eq:cRGD-M} (\textsc{cRGD-M}), OR \\
\eqref{Eq:Proj118-7} (\textsc{RGD-RMSProp}), OR \\
\eqref{Eq:Proj118-9} (\textsc{RGD-AdaDelta}); 
\STATE $ t \leftarrow t + 1$
\ENDIF
\end{algorithmic}
\end{algorithm}

\section{Experiment}\label{Sec:4}
In this section, we investigate the performance of our proposed Gaussian variational approximation methods. There are eight of them: the Gaussian variational approximation with Stiefel manifold in Section \ref{Sec:2.2}
and the Gaussian variational approximation with Grassmann manifold in Section \ref{Sec:2.3}, each is estimated with four manifold learning rules described in Section \ref{Sec:3.2}.
We start from Section~\ref{Sec:4.1} with a toy experiment where the data is generated from a non-linear regression model with outliers. We evaluate all eight proposed methods' sensitivity to the different numbers of factors and report the fitting performance as well as the convergence speed. We also show the infeasibility of the naive low-rank representation structure mentioned in Section~\ref{Sec:2.1}. In Section~\ref{Sec:4.2} and \ref{Sec:4.3}, we demonstrate the methods' prediction power with low-dimensional ($m < n$) and high-dimensional ($m \gg n$) datasets, respectively. The accuracy is compared against three baseline methods, including \textsc{VAFC} \citep{ong2018gaussian}, \textsc{SLANG} \citep{mishkin2018slang} and the conventional mean-field variational inference method.  SLANG is a variational inference algorithm that uses the natural gradient and the mean-field variational inference method is a special case of our framework with the factor number $p = 0$. All experiments are implemented in Matlab on a laptop with an Intel Core 64-bit i5-6600 3.3GHz CPU and 8G RAM.

\subsection{Simulation Study}\label{Sec:4.1}
We follow \cite{ong2018gaussian} and generate data to investigate our methods' performance. We are specifically interested in their convergence speed, fitting performance, and their sensitivity to different numbers of factors $p$. 

\begin{figure}
\centering
\includegraphics[width = \textwidth, trim = {0cm 0.1cm 0cm 0.6cm}, clip]{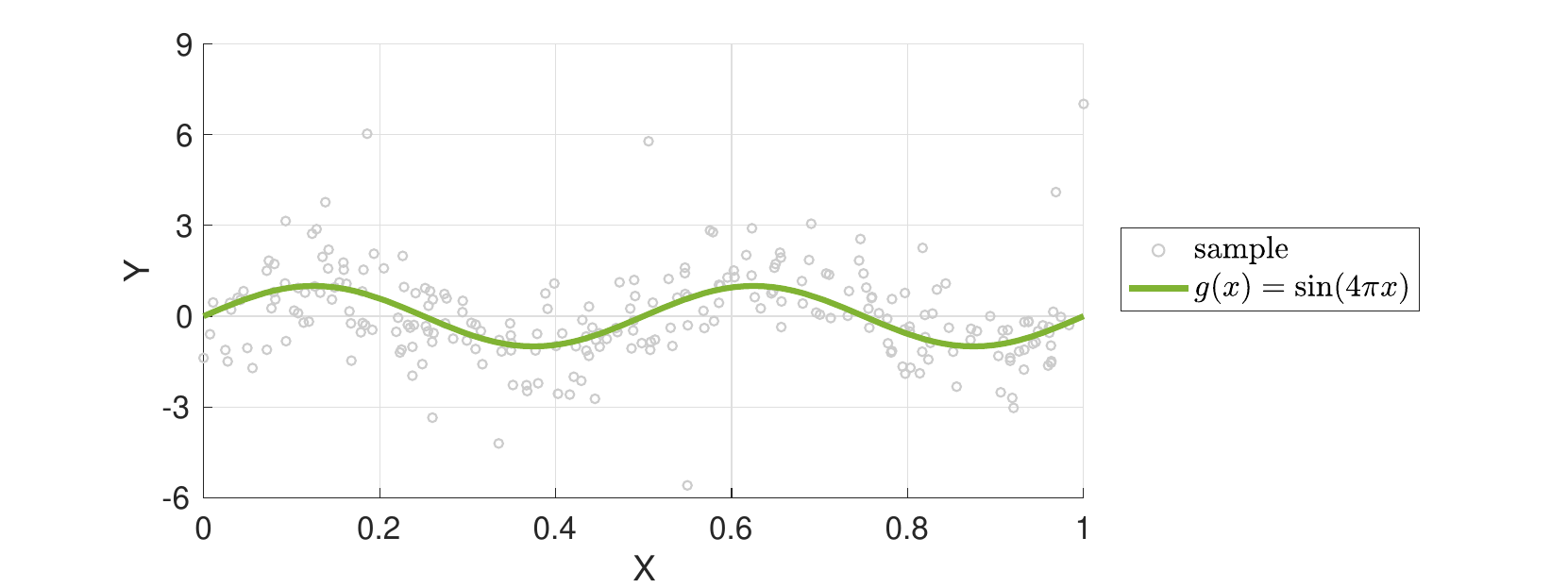}
\caption{The true function of the simulation experiment.}
\label{fig-e1}
\end{figure}

Consider a regression model
$$y_i = g(\mathbf{x}_i) + \epsilon_i,\;\;i=1,...,n=250,$$
where $g(\mathbf{x})=\sin(4\pi \mathbf{x})$ is the underlying function to be approximated. 
We generate the $\mathbf{x}_i$ from $U(0,1)$ and $\epsilon_i$ from the mixture
$$p(\epsilon_i) = 0.95\mathcal{N}(\epsilon_i; 0, \sigma^2) + 0.05\mathcal{N}(\epsilon_i; 0, a\sigma^2).$$
We set $\sigma^2 = 1$ and $a = 5$. The parameter $a$ controls the occurrence of extreme values. The methods' degree of robustness can then be justified by setting different $a$s. A visualization of the underlying function and the generated 250 samples is given in Figure~\ref{fig-e1}.

To approximate the underlying function $g$, we consider the \emph{penalized B-spline} (P-spline) regression model \citep{lang2004bayesian}. Denote by $b(\mathbf{x})$ a basis of P-splines with $k$ basis functions $b(\mathbf{x}) = (b_1(\mathbf{x}), b_2(\mathbf{x}), ..., b_k(\mathbf{x}))^{\top}$, and $\boldsymbol{\beta} = (\beta_1, \beta_2, ..., \beta_k)^{\top}$ the corresponding $k$ coefficients. We use $k=25$ in this example. The B-spline regression model is
$$
{y} = \boldsymbol{\beta}^{\top} b(\mathbf{x}) + \boldsymbol{\epsilon}.
$$
The model has a total number of $k+3$ random variables to infer. Aside from the aforementioned $k$ coefficients $\boldsymbol{\beta}$, there are three additional parameters $(\tau^2, \psi, \sigma^2)$ for describing the density of noise $\boldsymbol{\epsilon}$. Below we give full details to construct them under the Bayesian learning framework.

\cite{ong2018gaussian} use the likelihood
\[
p(y_i|\mathbf x_i, \boldsymbol{\beta})=0.95\mathcal{N}(\boldsymbol{\beta}^{\top} b(\mathbf{x}), \sigma^2) + 0.05 \mathcal{N}(\boldsymbol{\beta}^{\top} b(\mathbf{x}), 10\sigma^2)
\]
and the following prior for $\beta$ 
$$\beta | \tau^{2}, \psi \sim N\left(0,\left(\tau^{2}\right)^{-1} P(\psi)^{-1}\right)$$
with $P(\psi)^{-1}$ the matrix formed by the AR(1) model with the persistence coefficient $\psi$. 
The prior of the hyperparameters $\tau^{2}, \psi$ is
$$p(\tau^{2}, \psi)=p(1/ \tau^2) \times(1 / 0.99) I(0<\psi<0.99),$$
where $p(1/ \tau^2)$ is the Weibull density
$$
p(1 / \tau^2)= 
\begin{cases}
    \frac{a_\tau}{b_\tau}\left(\frac{1}{\tau^2 b_\tau}\right)^{a_\tau - 1}e^{-(1/\tau^2b_\tau)^{a_\tau}},& \text{if } 1 / \tau^2 > 1 \\
    0,  & \text{otherwise}
\end{cases}.
$$
Finally, the prior for $\sigma^2$ is an inverse Gamma distribution with shape parameter $\alpha_0$ and scale parameter $\beta_0$
$$P(\sigma^{2} ; \alpha_0, \beta_0) = \frac{\beta_{0}^{\alpha_{0}}}{\Gamma(\alpha_{0})}(\frac{1}{\sigma^{2}})^{\alpha_{0}+1}\exp(-\frac{\beta_{0}}{\sigma^{2}}).$$

\begin{figure*}[tbh]
\centering
\subfloat{\includegraphics[width = \textwidth, trim = 5.8cm 0.05cm 5.8cm 0.25cm, clip]{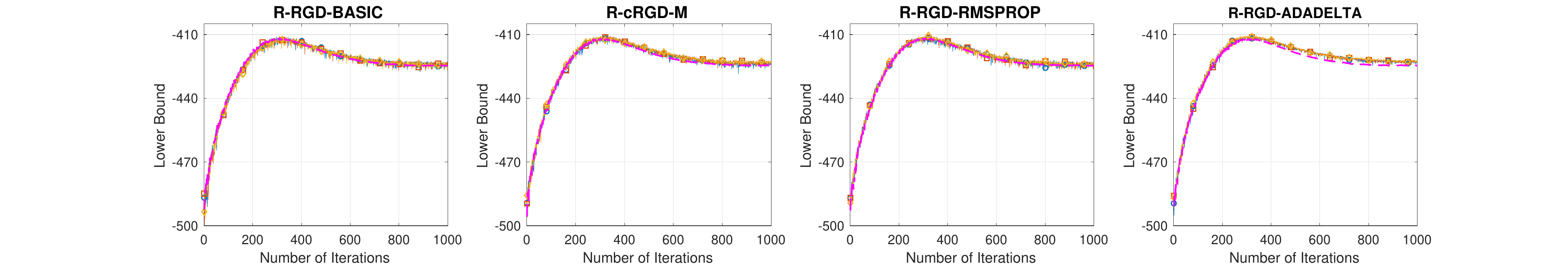}}\\
\subfloat{\includegraphics[width = \textwidth, trim = 5.8cm 0.05cm 5.8cm 0.25cm, clip]{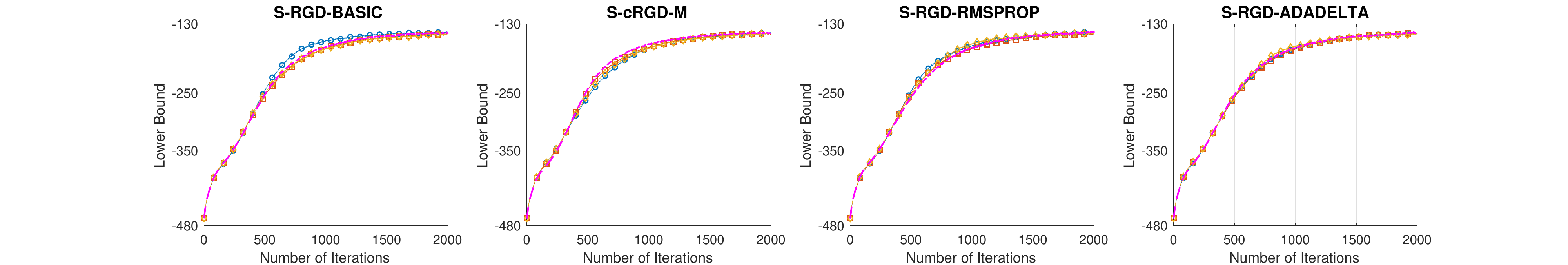}}\\
\subfloat{\includegraphics[width = \textwidth, trim = 5.8cm 0.05cm 5.8cm 0.25cm, clip]{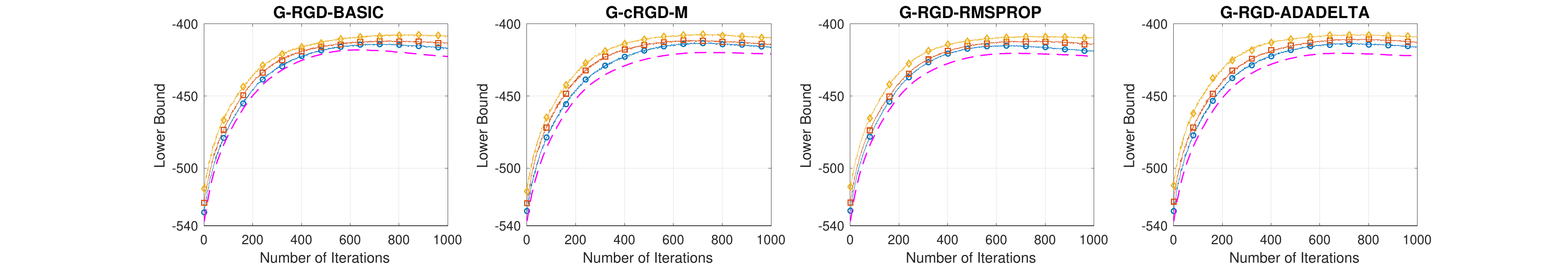}}\\
\subfloat{\includegraphics[clip,trim = 15cm 1.8cm 15cm 17cm, width=\textwidth]{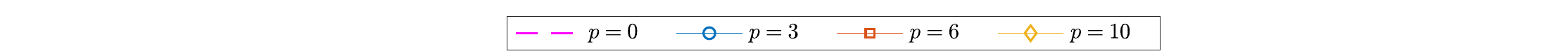}}\\
\caption{Lower bound approximation of the 12 methods with $p = 0, 3, 6$ and 10 factors. We show the Stiefel results in 2000 iterations, and Fixed-rank and Grassmann results in 1000 iterations.} 
\label{fig-1a}
\end{figure*}

\begin{figure*}[tbh]
\centering
\subfloat{\includegraphics[clip,trim = 5.8cm 0.05cm 5.8cm 0.25cm, width=\textwidth]{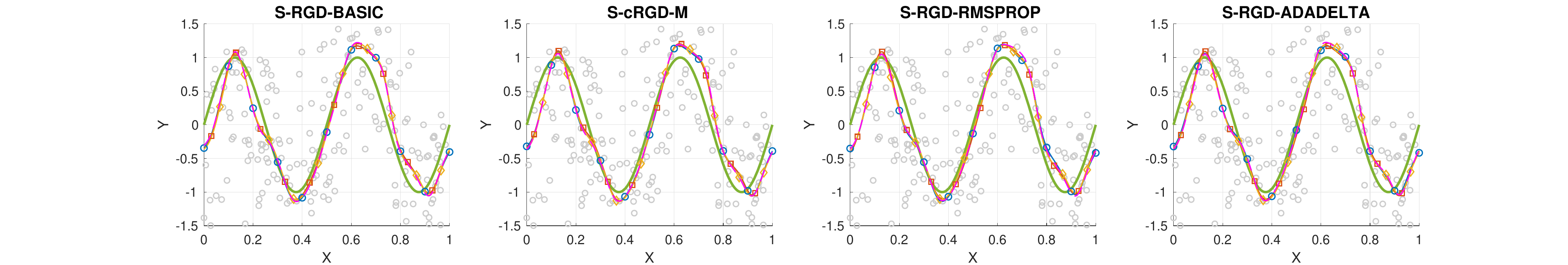}}\\
\subfloat{\includegraphics[clip,trim = 5.8cm 0.05cm 5.8cm 0.25cm,, width=\textwidth]{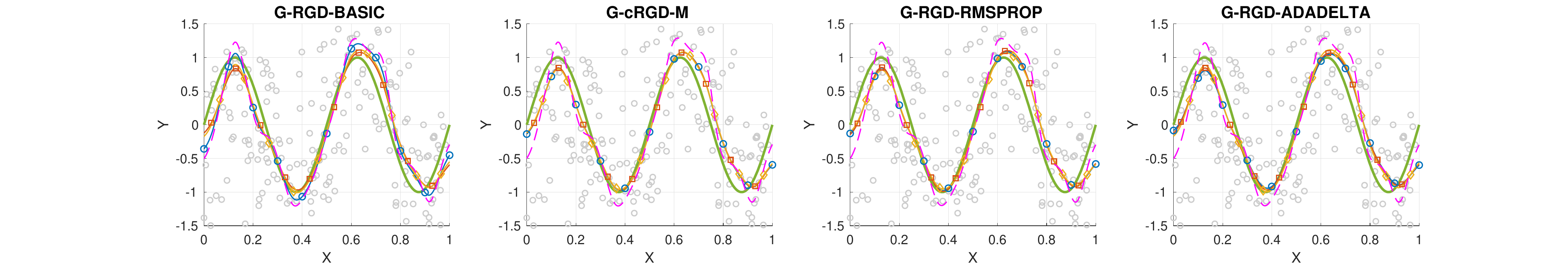}}\\
\subfloat{\includegraphics[clip,trim = 15cm 1.2cm 15cm 11cm, width=\textwidth]{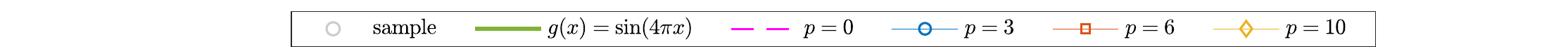}}\\
\caption{Fitting performance of the 8 Gaussian variational approximation methods with $p = 0, 3, 6$ and $10$ factors, estimated after 1000 iterations. The true function is $g(x) = \sin(4\pi x)$ (shown in green). The grey dots are the 250 sample data points within range [-1.5, 1.5]. The titles describe the method used.} 
\label{fig-1b}
\end{figure*}

The set of model parameters is $\boldsymbol{\theta} = \{\boldsymbol{\beta},\tau^2, \psi,\sigma^2\}$.
The Gaussian variational approximation methods for the posterior of $\boldsymbol{\theta}$ are obtained by the procedures outlined in Algorithm~\ref{Alg:1}. 
We are interested in their convergence speed, accuracy, and their sensitivity to different numbers of factors $p$.  The factorization of $\boldsymbol{\Sigma}$ follows \eqref{Eq:Proj118-2} for Stiefel manifold constraints, and \eqref{Eq:Sigma} for Grassmann manifold constraints. For the optimizers, we specify the learning rates $\alpha=0.05, \beta = 0.95$ and $\epsilon=10^{-6}$. 

The lower bounds of the proposed methods are visualized in Figure~\ref{fig-1a} with $p = 0, 3, 6$ and $10$ factors. We only plot the first 1000 iterations for the fixed-rank and Grassmann methods and 2000 iterations for the methods with Stiefel manifold, as it takes more iterations to converge. All values are averaged from 10 repetitions.

The first row shows the results from the Fixed-rank geometry methods. Their failure to converge properly supports the claims in Remark~\ref{Remark1}. In the second row, the four Stiefel manifold-based methods share a similar convergence speed. The selection of factor $p$ does not dramatically influence the convergence speed, as the lower bounds for the different $p$ choices do not significantly differ from each other. Also, including extra covariance information does not benefit the model approximation as expected. As displayed in the subplots, not all the lower bound lines are above the magenta dashed line (factor $p = 0$). In other words, more complex factor structures do not always outperform the naive independent covariance structure when implementing the Stiefel manifold method. The third row shows the lower bounds from the Grassmann manifold method, which tells a different story.
The lines are generally smoother and stabler than the ones in the second row, and attain the maximum in the first 1000 iterations.
In addition, the approximation process of the Grassmann methods can benefit from the  covariance structure. For all four methods, selecting a larger factor of $p$ always results in a better lower bound curve. 
The estimation results of the proposed Gaussian variational approximation methods are evaluated in Figure.~\ref{fig-1b}. 

All 32 regression models (with 4 factor values and 8 methods) capture the underlying function $g(x)$ reasonably well. The results with $p>0$ factors are generally less sensitive to the extreme values than the baseline mean-field methods with a diagonal covariance ($p = 0$). Since the samples are distributed less dense at $[0, 0.2]$ than at $[0.5, 0.7]$, the estimated lines are slimmer than the ground truth (green line) in the former region but wider in the latter region. Also, within $[0.5, 0.7]$ there are more observations above the green line with considerably large values, see Figure~\ref{fig-e1}. Consequently, higher peaks are approximated than the true function. The four Gaussian variational approximation methods with the Grassmann manifold constraint are more robust to extreme values than the ones with the Stiefel manifold. The fitted lines around stationary points are more similar to the ground truth. Also, the factorized methods outperform the baseline mean-field method, especially when working around the outliers. In terms of the four optimizers, they obtain similar performance with Stiefel and Grassmann geometry. We will leave further comparison and discussion in Section~\ref{Sec:4.2} and \ref{Sec:4.3} with complex data structure.

Based on what we have observed from these toy experiments, we conclude that both the Stiefel and Grassmann manifolds provide accurate results at a considerably fast speed, while the conventional Fixed-rank geometry fails to handle the constrained optimization. In addition, the Grassmann methods outperform the Stiefel methods in terms of both convergence rates and approximation accuracy. 

In the next two sections, we will consider empirical datasets including both low-dimensional and high-dimensional settings to examine our Gaussian variational approximation methods with more complex structures. 

\begin{table}
\caption{The average training error and test error rates for the ionosphere data with five-fold cross-validation and factor $p = 4$.
}\label{Table2}
\begin{center}\begin{tabular}{lrrr}
\hline
\textbf{Method} & \textbf{Training Error}(\%) & \textbf{Test Error}(\%)  & \textbf{Time}(sec) \\ \hline
\textsc{VAFC} ($p=0$)    & 0.64   & 7.65 & \textbf{12.51} \\
\textsc{VAFC}            & \textbf{0.36} & 8.22 & 23.04 \\
\textsc{SLANG}           & 7.55   & 14.20& 190.93 \\
\hline 
\textsc{S-RGD-Basic}    & \textbf{0.36} & 7.65 & 25.02 \\      
\textsc{S-cRGD-M}        & 0.43   & 7.65 & 24.71 \\        
\textsc{S-RGD-RMSProp}   & 0.43   & \textbf{7.09} & 25.29 \\        
\textsc{S-RGD-AdaDelta}  & 0.43   & 7.37 & 27.27 \\\hline 
\textsc{G-RGD-Basic}     & 1.00   & 7.38 & 24.78 \\    
\textsc{G-cRGD-M}        & 0.71   & 7.38 & 24.75 \\   
\textsc{G-RGD-RMSProp}   & 0.71   & \textbf{7.09} & 24.43 \\  
\textsc{G-RGD-AdaDelta}  & 0.64   & \textbf{7.09} & 25.59 \\ \hline
\end{tabular}
\end{center}
\end{table}

\begin{figure}
\centering
\subfloat{\includegraphics[width = \textwidth, trim = 5.8cm 0.05cm 5.8cm 0.25cm, clip]{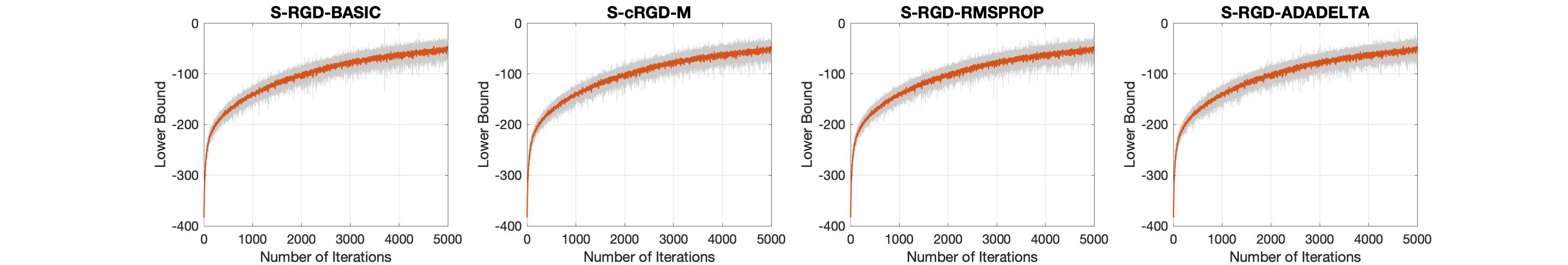}}\\
\subfloat{\includegraphics[width = \textwidth, trim = 5.8cm 0.05cm 5.8cm 0.25cm, clip]{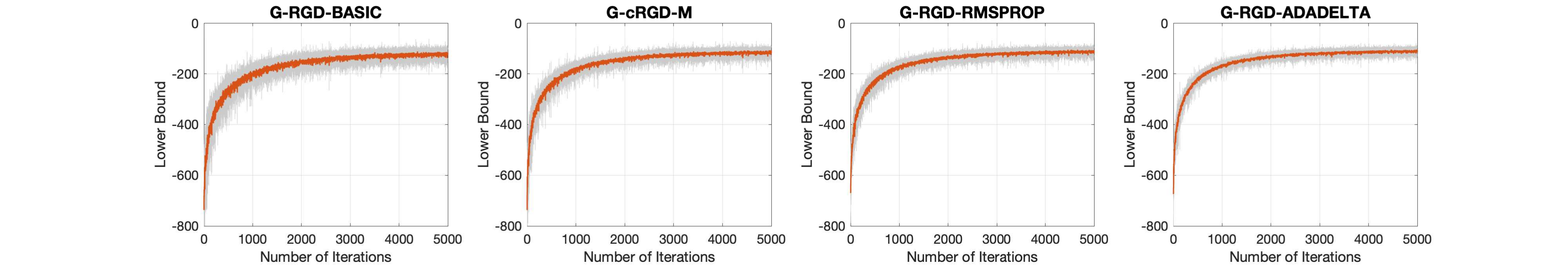}}\\
\caption{Lower bound approximation of the 8 methods with $p = 4$ factors in 5000 iterations. The convergence traces (in orange) are averaged lower bounds over 30 independent repetitions. The area (in grey) indicate the variation within one standard deviation.}
\label{fig-2}
\end{figure}

\subsection{Low Dimensional Predictive Inference}\label{Sec:4.2}
We first evaluate our methods on the logistic regression for a low dimensional binary classification problem. We select the \textbf{ionosphere} data from the UCI Machine Learning Repository \citep{Dua:2019} with 351 samples. The task is to classify 225 positive signals that could identify an ionosphere structure and 126 negative instances that are insensitive to the free electrons. We follow the same data processing operations as \cite{ong2018gaussian} and expand the attributes from 34 to 111. Given $\widehat{\mathbf x}_i\in\mathbb{R}^q$, the likelihood of the response $y_i\in\{-1, 1\}$ is
$$
p(y_i|\mathbf x_i, \boldsymbol{\beta}) = \frac1{(1+\exp\{-y_i \boldsymbol{\beta}^{\top}\mathbf x_i\})},
$$ 
where $\mathbf x_i= [1, \widehat{\mathbf x}^{\top}_i]^{\top}$ and $\boldsymbol{\beta}$ denotes the coefficient vector. We use the Gaussian prior for $\boldsymbol{\beta}$. The prediction performance of five-fold cross-validation is recorded in Table~\ref{Table2}. The best results are highlighted in terms of the training error, test error and training time.

The overall accuracy of our proposed methods improves slightly from the two benchmarks \textsc{VAFC} \citep{ong2018gaussian} and mean-field models, but outperforms \textsc{SLANG} \citep{mishkin2018slang} significantly. All eight proposed methods have a lower test error, where around 5 samples are misclassified among 70 test instances. In particular, the Stiefel methods obtain both lower training error and test error, while the Grassmann methods further cut down the test error with a higher training error. Therefore, we conclude that the imposed manifold constraints, especially the Grassmann manifold constraint, make the methods less likely to overfit the data. The computational times of all eight methods are similar and similar to that of VAFC, and much less than that of SLANG.
In terms of the optimizers, the two new methods (\textsc{RGD-RMSProp} and \textsc{RGD-AdaDelta}) outperforms the two conventional update rules (\textsc{RGD-Basic} and \textsc{cRGD-M}) in both Stiefel and Grassmann settings. In particular, the \textsc{RGD-RMSProp} methods obtain the lowest test error with a fast speed.

Figure~\ref{fig-2} compares the estimation of lower bounds for the 8 proposed methods with $p = 4$ factors over $5000$ iterations. Similar to the toy example, the shape of each curve depends on the manifold constraints. All the 8 models shrink their lower bounds quickly in the first $2000$ iterations. However, it takes longer for Stiefel methods to flatten the curve, and the oscillation is more significant than the Grassmann methods along the update process. The reason is that the Stiefel methods have an extra scaling factor $\mathbf{D}_1$ to update while the Grassmann methods only take care of $\mathbf B$ and $\mathbf{D}_2$. For different optimizers on the same manifold constraints, the latter two optimizers (\textsc{RGD-RMSProp} and \textsc{RGD-AdaDelta}) update slightly faster than the former two (\textsc{RGD-Basic} and \textsc{cRGD-M}), but the advantage does not distinguish clearly.

\subsection{High Dimensional Predictive Inference}\label{Sec:4.3}
In the last experiment, we consider the binary \textbf{Leukemia Cancer} dataset with $7120$ predictors available at \url{http://www.ntu.edu.sg/home/elhchen/data.htm}. This classification problem contains 38 training instances and 34 testing instances. We evaluate the methods' scalability with $m\gg n$. The Horseshoe prior \citep{carvalho2010horseshoe} is applied due to the sparse data structure. The predictive accuracy is reported in Table~\ref{Table3}. The performance of the 8 methods is compared against the mean-field variational approximation and \textsc{VAFC}. We don't report the results from \textsc{SLANG} as it fails to converge after 96 hours.

\begin{table}
\begin{center}
\caption{Average training error, test error and running time for the Leukemia Cancer data with $p = 4$ and iteration$=5000$.}\label{Table3}
\begin{tabular}{lrrr}
\hline
\textbf{Method} & \textbf{Training Error} & \textbf{Test Error} & \textbf{Time}(sec) \\ 
\hline
\textsc{VAFC} ($p=0$)    & 0/38  & 7/34  & \textbf{64.08}       \\
\textsc{VAFC}            & 0/38  & 3/34  & 8021.71       \\
\textsc{SLANG}           & -     & -     & -        \\
\hline
\textsc{S-RGD-Basic}     & 0/38  & \textbf{2/34} & 10013.90      \\
\textsc{S-cRGD-M}        & 0/38  & \textbf{2/34} & 10031.65      \\
\textsc{S-RGD-RMSProp}   & 0/38  & \textbf{2/34} & 10011.27      \\
\textsc{S-RGD-AdaDelta}  & 0/38  & \textbf{2/34} & 10005.28      \\
\hline
\textsc{G-RGD-Basic}     & 0/38  & \textbf{2/34} & \textbf{6482.94}       \\
\textsc{G-cRGD-M}        & 0/38  & \textbf{2/34} & \textbf{6438.22}       \\
\textsc{G-RGD-RMSProp}   & 0/38  & 3/34          & \textbf{6590.98}       \\
\textsc{G-RGD-AdaDelta}  & 0/38  & \textbf{2/34} & \textbf{6202.85}       \\ 
\hline
\end{tabular}
\end{center}
\end{table}

As the sample size is relatively small, the proposed methods achieve similar performance in terms of predictive accuracy. Still, they all beat the two benchmark models. Specifically, the two negative samples are identically misclassified among all methods, which are wrongly predicted as positive. One of the possible reasons is that the training dataset is unbalanced with 27 positive samples against 11 negative samples. In terms of the operating speed, the conventional mean-field method stands out with its trivial structure. However, the fast computation sacrifices model accuracy, where the test error is significantly higher than the other methods. For the rest of the models, the Grassmann models are faster than other structures. In terms of the update methods, the \textsc{RGD-AdaDelta} scheme is faster than other optimization methods.

\section{Discussion and Conclusion}\label{Sec:5}
This paper proposes two new manifolds-assisted optimization methods to resolve the identification issue of matrix factorization in the Gaussian variational approximation. We also develop two adaptive Riemannian gradient descent schemes, \textsc{RGD-RMSProp} and \textsc{RGD-AdaDelta}, to accelerate the optimization process. The proposed methods are assessed in both low- and high-dimensional situations. The experiments demonstrate the better performance of the proposed methods than the benchmark mean-field, \textsc{VAFC}, and \textsc{SLANG} models in terms of prediction accuracy and convergence speed.

It should be aware that the overhead cost of optimization on manifolds is usually higher than its Euclidean counterpart. 
This is because of the extra calculation of the retraction map and vector transport.
The current research restricts the distribution of the latent variables to the multivariate Gaussian. Extension to other variational distributions, such as the exponential family, would be a potential research direction.

\bibliography{JCGS}

\end{document}